\documentclass[runningheads]{llncs}
\usepackage[T1]{fontenc}
\usepackage{makecell}
\usepackage{graphicx}
\usepackage{subcaption}
\usepackage{float} 
\usepackage{booktabs} 
\usepackage{multirow}   
\usepackage{amsmath}
\usepackage{mathrsfs}
\usepackage{url}
\usepackage{amsfonts}

\begin{document}
\title{Disentanglement and Assessment of Shortcuts in Ophthalmological Retinal Imaging Exams}
\titlerunning{Disentanglement in Ophthalmological Retinal Imaging Exams}
%
\author{Leonor Fernandes\inst{1,2}
\and
Tiago Gonçalves\inst{1,2}
\and
João Matos\inst{3}
\and
Luis Nakayama\inst{4}
\and
Jaime S. Cardoso\inst{1,2}
}

%
\authorrunning{L. Fernandes et al.}
%
\institute{Faculdade de Engenharia, Universidade do Porto, Portugal 
\and 
Institute for Systems and Computer Engineering, Technology and Science, Porto, Portugal
\and 
University of Oxford, United Kingdom
\and 
Federal University of São Paulo, São Paulo, Brasil
}


%
\maketitle              
\begin{abstract}
Diabetic retinopathy (DR) is a leading cause of vision loss in working-age adults. While screening reduces the risk of blindness, traditional imaging is often costly and inaccessible. Artificial intelligence (AI) algorithms present a scalable diagnostic solution, but concerns regarding fairness and generalization persist.
This work evaluates the fairness and performance of image-trained models in DR prediction, as well as the impact of disentanglement as a bias mitigation technique, using the diverse mBRSET fundus dataset. Three models, ConvNeXt V2, DINOv2, and Swin V2, were trained on macula images to predict DR and sensitive attributes (SAs) (e.g., age and gender/sex). Fairness was assessed between subgroups of SAs, and disentanglement was applied to reduce bias.
All models achieved high DR prediction performance in diagnosing (up to 94\% AUROC) and could reasonably predict age and gender/sex (91\% and 77\% AUROC, respectively). Fairness assessment suggests disparities, such as a 10\% AUROC gap between age groups in DINOv2. Disentangling SAs from DR prediction had varying results, depending on the model selected. Disentanglement improved DINOv2 performance (2\% AUROC gain), but led to performance drops in ConvNeXt V2 and Swin V2 (7\% and 3\%, respectively).
These findings highlight the complexity of disentangling fine-grained features in fundus imaging and emphasize the importance of fairness in medical imaging AI to ensure equitable and reliable healthcare solutions.


\end{abstract}
\section{Introduction}
Diabetic retinopathy (DR) is one of the most common complications of diabetes and a leading cause of visual impairment and blindness worldwide. The disease develops when chronically high blood glucose levels damage the small blood vessels that supply the retina, resulting in swelling, leakage or bleeding. In response, the eye may attempt to grow new, but often abnormal and fragile blood vessels that can further impair vision or lead to retinal detachment. In its early stages, the condition is typically asymptomatic. As DR advances, individuals may experience symptoms such as blurry vision and floating spots in their vision. If left untreated, DR can lead to irreversible blindness. Regular eye examinations are essential for people with diabetes, as early detection and timely treatment can slow or prevent the progression of DR and help preserve vision \cite{ch2:treatment}.

In this work, we tackled the role of disentanglement in mitigating fairness issues in DR prediction in retinal fundus images, making the following contributions:
\begin{enumerate}
    \item Analysis of mBRSET, a handheld fundus camera dataset, for group fairness across sensitive attribute (SA) subgroups, using the area under the receiver operating characteristic curve (AUROC), decision curve analysis, and risk distribution plots;
    \item Development of a disentangled autoencoder architecture to separate medically relevant features from SA-related information; 
    \item Test of the architecture as a fairness mitigation technique, by comparing its performance and fairness outcomes against baseline models.
\end{enumerate}
The code related to the implementation of this paper is publicly available in a GitHub repository\footnote{\url{https://github.com/public_if_accepted}}.


\section{Background and Related Work}
DR datasets are often imbalanced, which can exacerbate shortcut learning, as models may learn dataset-specific associations that do not generalize well to new populations or clinical settings \cite{ch3:shortcut_generalization}.
In medical imaging, SAs are often encoded in the data, allowing models to rely on these features as predictive shortcuts, which can lead to both unfairness and decreased domain generalization \cite{ch3:shortcut_fairness_limits}.
To address these challenges, various bias mitigation techniques have been proposed, including disentanglement learning, adversarial debiasing, and integration of fairness-related penalties \cite{ch3:bias_mitigation_review}. Adding fairness constraints or regularization terms to the loss function is a common approach. However, applying this approach to deep neural networks (DNNs) is challenging, due to their overparameterized nature, which may limit fairness generalization \cite{ch3:fairness_constraints}.

Disentanglement learning separates the underlying independent factors of variation within data, such as medical features and SAs. This approach can improve predictive performance \cite{ch3:disen_predictive}, improve fairness \cite{ch3:FairDisCo}, and mitigate shortcut learning \cite{ch3:disen_fair_xray}. In ophthalmology, disentanglement has been applied to tasks like anonymization and de-identification of retinal images \cite{ch3:anonym_helena,ch3:anonym_zhao}, as well as domain generalization across imaging devices and clinical sites \cite{ch3:DG_VAE,ch3:DG_loss,ch3:DG_DECO}. While its potential to improve fairness has been explored in other medical imaging domains \cite{ch3:FairDisCo,ch3:disen_fair_xray}, it remains underexplored in ophthalmology.

\section{Methodologies}

\subsection{Dataset and Preprocessing}
We used the mBRSET dataset, introduced in~\cite{ch4:mBRSET_}, the first publicly available dataset of retinal images captured using handheld retinal cameras. It contains 5,164 images from 1,291 patients of diverse ethnic backgrounds in Brazil and includes demographic information such as sex, age, and insurance. We refer the reader to~\cite{ch4:mBRSET_} for an in-depth analysis of the remaining attributes. 

For the binary classification task, DR severity was binarized using the ICDR scale: \textbf{Normal} (levels 0-1) and \textbf{Referable} (levels 2-4). Only high-quality, macula-centered images were used to ensure anatomical consistency. All images were resized to $224\times 224$ pixels and normalized using ImageNet statistics (i.e., mean and standard deviation) to align with pretrained model expectations~\cite{ch4:ImageNet}. The dataset was split into training (70\%), validation (10\%), and testing (20\%) subsets, stratified by DR distribution and patient identity to prevent data leakage. The DR prevalence in each subset was 18\%, 30\%, and 17\%, respectively. Data augmentation (i.e., random rotation, flips, color jitter, Gaussian blur) was applied during training to improve generalization.

\subsection{Models and Training}
We trained three image-based deep learning architectures, including a convolutional neural network, ConvNeXt V2~\cite{ch4:convnext}, and vision transformer-based models, DINOv2~\cite{ch4:dinov} and Swin V2~\cite{ch4:swin}. To address class imbalance, we employed focal loss, with class weights computed from the training distribution~\cite{ch4:focal_loss}. Models were trained using the Adam optimizer~\cite{ch4:adam_opt} with early stopping based on F1-score. Key hyperparameters included a hidden dimension of 128, a learning rate of \(1 \times 10^{-5}\), and a batch size of 4 for ConvNeXt V2 and DINOv2, and 2 for Swin V2.

\subsection{Fairness Assessment}
To evaluate fairness, we used a group fairness approach, focusing on AUROC, risk distribution plots, and decision curve analysis (DCA), stratified by groups of patients defined by SAs. The analysis considered five sensitive attributes: age ($\leq50$ vs. $>50$), sex (female vs. male), education (literate vs. illiterate), insurance (none vs. insured), and obesity (non-obese vs. obese).




\subsection{Disentanglement Architecture}
\begin{figure}[htbp]
    \centering
    \includegraphics[width=1\linewidth]{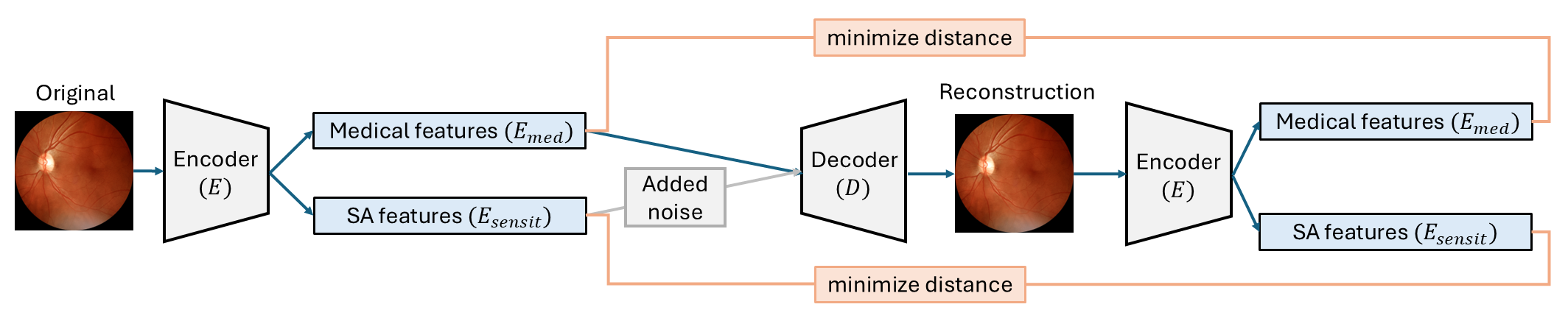}
    \caption{Overview of the disentanglement network.}
    \label{fig:disentanglement_network}
\end{figure}

To decrease model bias, we developed a disentanglement architecture that could separate the medical information from the information of a specified SA. This work is based on the architecture proposed in~\cite{ch3:anonym_helena}, which used a generative model for disentangling identity and medical characteristics in images. The adapted architecture is represented in Figure \ref{fig:disentanglement_network}. 

The four main components of the network are:
\begin{itemize}
    \item \textbf{Encoder $(E)$}: Transforms the input image into two independent vectors, $(E_{\mathrm{med}})$, that encodes the medical features, and $(E_{\mathrm{sensit}}$), that encodes the SA features.
    
    \item \textbf{Diabetic Retinopathy Classifier $(C_{\mathrm{med}})$}: Guides $E_{\mathrm{med}}$ to capture medically relevant information by classifying DR severity.
    
    \item \textbf{Sensitive Attribute Classifier $(C_{\mathrm{sensit}})$}: The SA classification task is used to approximate the $E_{\mathrm{sensit}}$ vector to the geometric space of the SA features.
    
    \item  \textbf{Decoder $(D)$}: Trained to reconstruct the original image from the vectors $E_{\mathrm{med}}$ and $E_{\mathrm{sensit}}$, promoting realistic reconstruction and enforcing the utility of both vectors.

\end{itemize}

The disentangled architecture was trained on the mBRSET dataset using a latent dimension of 256, a batch size of 32, a learning rate of \(5 \times 10^{-5}\), and a weight decay of \(1 \times 10^{-6}\).  The model is trained using a total loss function composed of three weighted components, each with a specific objective. The loss terms are:

\begin{itemize}
    \item  \textbf{Disentanglement Loss} $(\mathscr{L}_{\text {disent}})$: Represented in Equation \eqref{eq:disent_loss}, promotes independence between latent vectors $E_{\mathrm{med}}$ and $E_{\mathrm{sensit}}$ by penalizing scenarios where modifying a latent vector affects the other. Gaussian noise is added to only one of the latent vectors to simulate perturbations, and a new, altered image ($I_i$) is generated. 
    

        

\begin{equation}
\label{eq:disent_loss}
    \begin{array}{l}
\mathscr{L}_{\text {disent}}=\mathbb{E}_{I \sim p_{d}(I)}\left[\sum _ { i \in \{ \text {med,sensit} \} } \left(\left(E_{i}\left(I_{\text {ori}}\right)-E_{i}\left(I_{i}\right)\right)^{2}\right.\right. \\
\left.\left.+\sum_{\substack{j \in\{\text {med,sensit}\} \\
j \neq i}}\left(E_{j}\left(I_{\text {ori}}\right)-E_{j}\left(I_{i}\right)\right)^{2}\right)\right]
\end{array}
\end{equation}

    \item \textbf{Classification Loss} $(\mathscr{L}_{\text {classifier}})$: Ensures that $E_{\mathrm{med}}$ and $E_{\mathrm{sensit}}$ encode the intended features by evaluating their effectiveness on DR and SA classification tasks, respectively. $\mathscr{L}_{\text {classifier}}$ is represented in Equation \eqref{eq:classifier_loss}. 

\begin{equation}
\label{eq:classifier_loss}
    \begin{array}{l}
\mathscr{L}_{\text {classifier}}=-\lambda_{\text {med}} \sum_{c} y_{\text {med}}(c) \log \left(p_{\text {med}}(c)\right) \\
-\lambda_{sensit} \sum_{k} y_{sensit}(k) \log \left(p_{sensit}(k)\right)
\end{array}
\end{equation}

    \item \textbf{Realism Loss} $(\mathscr{L}_{\text {realism}})$: Promotes high-quality image reconstruction by optimizing pixel-level similarity. This loss combines the Structural Similarity Index Measure (SSIM) and Peak Signal-to-Noise Ratio (PSNR), normalized with a threshold parameter $\alpha = 48$ for PSNR. $\mathscr{L}_{\text {realism}}$ is represented in Equation \eqref{eq:realism_loss}.



\begin{equation}
\label{eq:realism_loss}
\mathscr{L}_{\text{realism}} = \mathbb{E}_{I \sim p_d(I)} \left[
    \left(1 - \operatorname{SSIM}\left(I_{\text{ori}}, D\left(E(I_{\text{ori}})\right)\right)\right) +
    \left(1 - \frac{1}{\alpha} \operatorname{PSNR}\left(I_{\text{ori}}, D\left(E(I_{\text{ori}})\right)\right)\right)
\right]
\end{equation}

    \item \textbf{Total Loss} $(\mathscr{L}_{\text {total}})$: Represented in Equation~\eqref{eq:total_loss}, combines the three loss terms into a total training objective. The relative contribution of the realism and disentanglement losses is controlled by the weighting factors $\lambda_r$ and $\lambda_d$, which were set to 1 and 5, respectively.

\end{itemize}

\begin{equation}
\label{eq:total_loss}
    \mathscr{L}_{\text {total }}=\mathscr{L}_{\text {classifier}}+\lambda_{r} \mathscr{L}_{\text {realism}}+\lambda_{d} \mathscr{L}_{\text {disent}}
\end{equation}

\section{Results and Discussion}
\subsection{Baseline Results}

\subsubsection{Classification of image models.}
Among the models, ConvNeXt V2 exhibited the highest overall performance, with an AUROC of 94\%, and DINOv2 consistently underperformed, achieving lower AUROC scores and net benefit, particularly in identifying referable DR cases. 
Models consistently performed better in identifying non-DR than referable DR. This is likely attributable to class imbalance, as referable DR cases only correspond to 17\% of mBRSET. 

\subsubsection{Sensitive attribute group prediction.}
 For the mBRSET dataset, the predicted SAs were age, sex, educational level, insurance, and obesity (see Table~\ref{tab:sa_pred}). Age was predicted well by all models, while sex prediction was moderate, with the exception of DINOv2, which performed poorly. All models struggled to predict educational level, insurance, and obesity.

\begin{table}[htbp]
    \caption{Performance of SA prediction using different models for the test set, reported as AUROC (\%) and balanced accuracy (BA) (\%).}
    \label{tab:sa_pred}
    \centering
    \resizebox{\textwidth}{!}{%
    \begin{tabular}{|c|cc|cc|cc|cc|cc|}
        \hline
        \multirow{2}{*}{\textbf{Model}}
        & \multicolumn{2}{c|}{\textbf{Age}} 
        & \multicolumn{2}{c|}{\textbf{Sex}} 
        & \multicolumn{2}{c|}{\textbf{Educational Level}} 
        & \multicolumn{2}{c|}{\textbf{Insurance}} 
        & \multicolumn{2}{c|}{\textbf{Obesity}} \\
        \cline{2-11}
        & \textbf{AUROC} & \textbf{BA} 
        & \textbf{AUROC} & \textbf{BA} 
        & \textbf{AUROC} & \textbf{BA} 
        & \textbf{AUROC} & \textbf{BA} 
        & \textbf{AUROC} & \textbf{BA} \\
        \hline
        ConvNeXt V2 & 90.50 & 69.09 & 76.60 & 68.29 & 56.96 & 53.57 & 46.65 & 50.00 & 63.14 & 50.00 \\
        DINOv2      & 90.73 & 77.32 & 53.35 & 50.00 & 57.01 & 50.00 & 48.79 & 50.00 & 40.29 & 50.00 \\
        Swin V2     & 87.85 & 78.64 & 76.79 & 69.89 & 63.02 & 52.64 & 41.74 & 50.00 & 58.50 & 50.00 \\
        \hline
    \end{tabular}
    }
\end{table}

\subsubsection{Fairness analysis of DR prediction with mBRSET.} Table~\ref{tab:auroc_performance_augmented} shows AUROC disparities across models and SA groups.
ConvNeXt V2 had the most balanced performance, with AUROC differences typically within 1-3\%, indicating strong generalizability and fairness.
Swin V2 also performed consistently, with most AUROC disparities remaining under 4\%.  It exhibited slightly higher discrepancies, particularly between different educational levels and obesity groups. 
DINOv2 had the largest discrepancies, including age (10\%), educational level (7\%), and insurance (12\%), raising fairness concerns.


\begin{table}[htbp]
    \caption{Performance of DR prediction across SA groups using different models for the test set, reported as AUROC (\%). N is the number of observations in the subgroup.}
    \label{tab:auroc_performance_augmented}
    \centering
    \begin{tabular}{|c|c|c|c|c|c|}
        \hline
        \textbf{\makecell{Attributes \\
        of mBRSET}} & \textbf{Subgroups} 
        & \textbf{N}  
        & \textbf{ConvNeXt V2} 
        & \textbf{DINOv2} 
        & \textbf{Swin V2} 
        \\
        \hline
        \multirow{1}{*}{Full} &  -& 498 & 94.33 & 88.31 & 90.96 \\
        \hline
        \multirow{2}{*}{Age} 
            & Group0 (below or equal to 50) & 81 & 97.06 & 80.09 & 88.69 \\
            \cline{2-6}

            & Group1 (above 50)& 417 & 93.79 & 89.81 & 91.41 \\
        \hline
        \multirow{2}{*}{Sex} 
            & Group0 (female)& 329 & 95.22 & 88.95 & 91.10 \\
            \cline{2-6}

            & Group1 (male)& 169 & 93.43 & 88.21 & 90.83 \\
        \hline
        \multirow{2}{*}{\makecell{Educational\\level}} 
            & Group0 (literate)& 222 & 95.22 & 92.43 & 92.74 \\
            \cline{2-6}
            & Group1 (illiterate)& 268 & 93.71 & 85.52 & 89.73 \\
        \hline
        \multirow{2}{*}{Insurance} 
            & Group0 (no insurance)& 455 & 94.26 & 88.86 & 91.27 \\
            \cline{2-6}
            & Group1 (insurance)& 39 & 97.14 & 77.14 & 85.71 \\
        \hline
        \multirow{2}{*}{Obesity} 
            & Group0 (no obesity)& 450 & 93.87 & 87.93 & 91.10 \\
            \cline{2-6}
            & Group1 (obesity)& 40 & 100.00 & 100.00 & 100.00 \\
        \hline
    \end{tabular}
\end{table}

\subsection{Disentanglement Architecture}
\begin{figure}[htbp]
    \centering

    \begin{subfigure}[t]{0.32\linewidth}
        \centering
        \includegraphics[width=\linewidth]{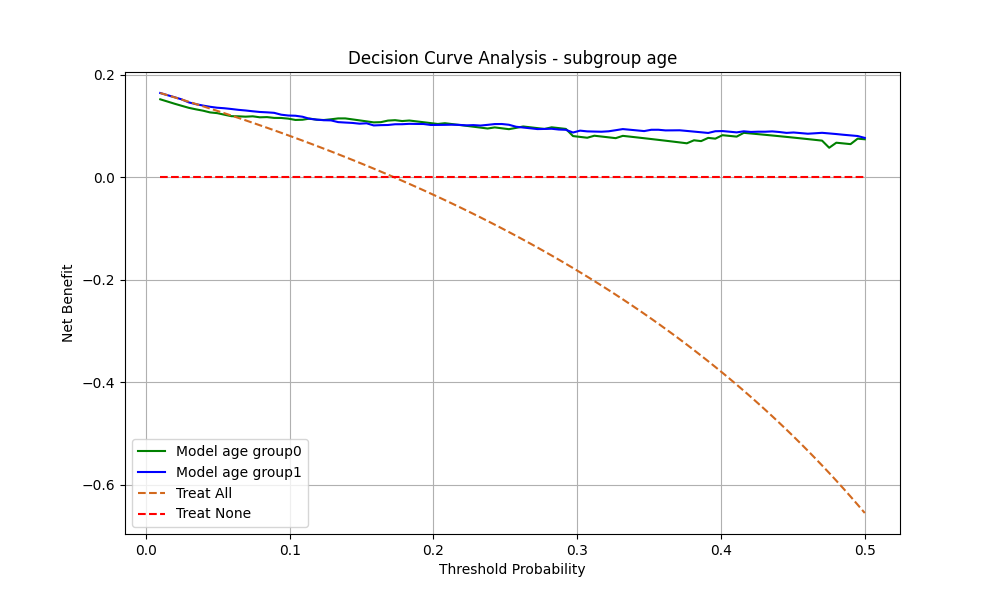}
        \includegraphics[width=\linewidth]{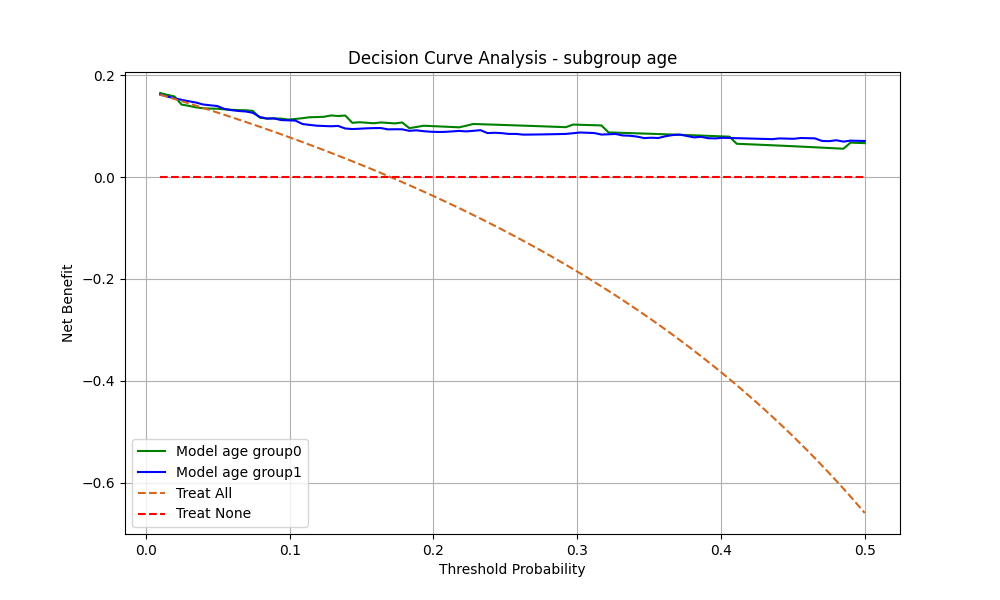}
        \caption{ConvNeXt V2 - Age (DCA).}
        \label{fig:mBRSET_conv_age_dca_combined}
    \end{subfigure}
    \hfill
    \begin{subfigure}[t]{0.32\linewidth}
        \centering
        \includegraphics[width=\linewidth]{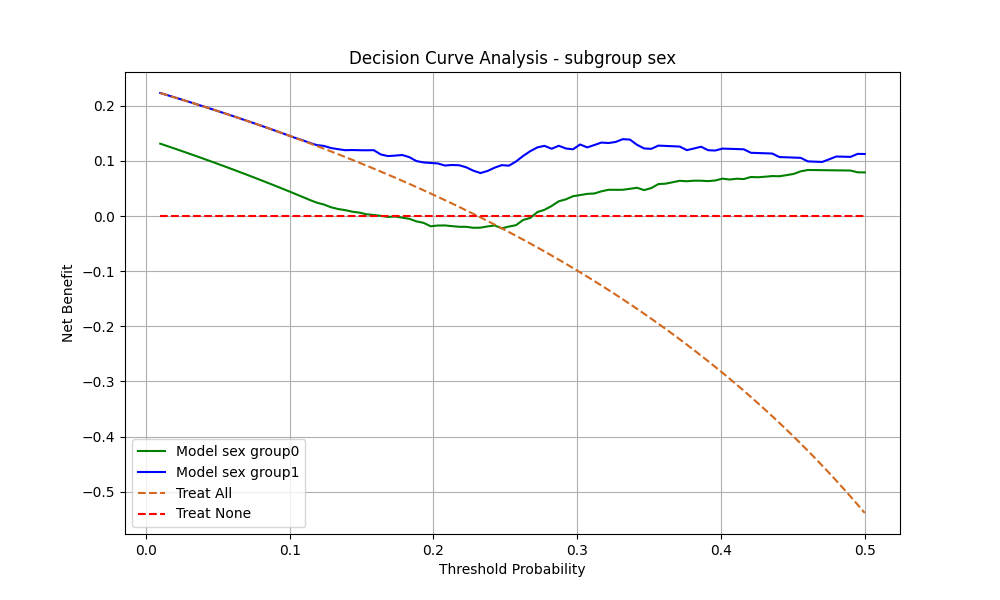}
        \includegraphics[width=\linewidth]{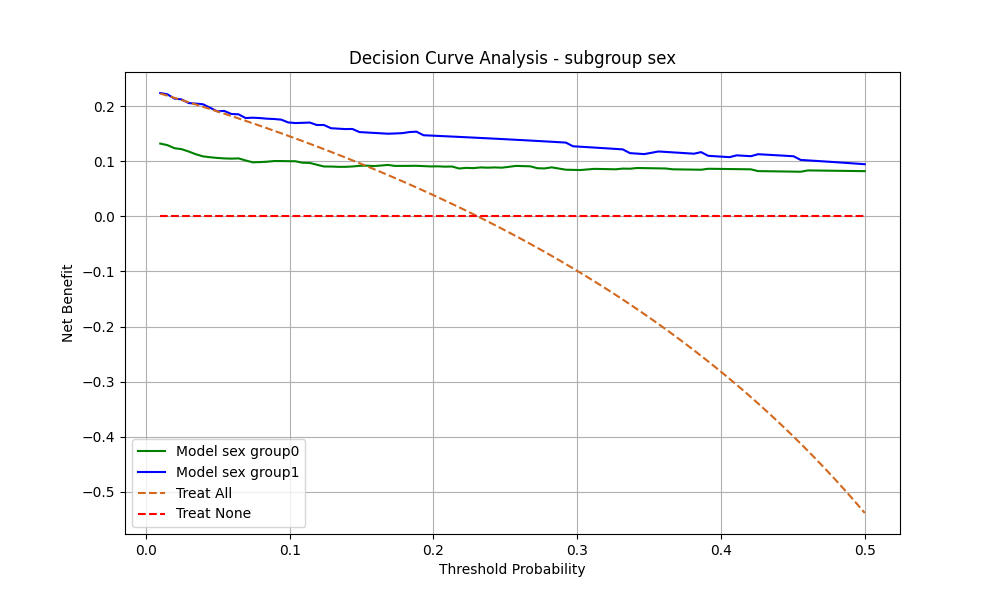}
        \caption{DINOv2 - Sex (DCA).}
        \label{fig:mBRSET_dino_sex_dca_combined}
    \end{subfigure}
    \hfill
    \begin{subfigure}[t]{0.32\linewidth}
        \centering
        \includegraphics[width=\linewidth]{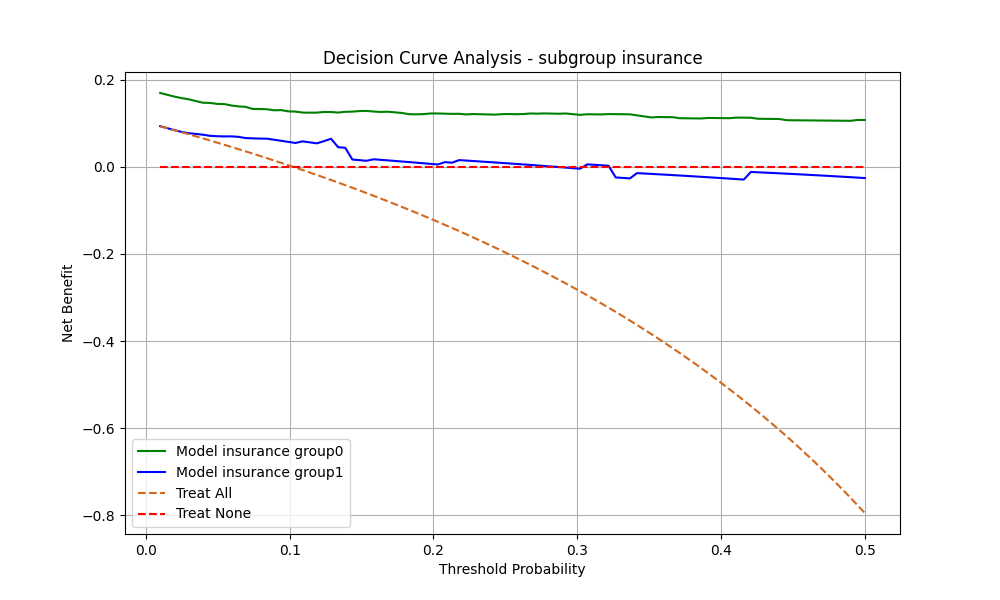}
        \includegraphics[width=\linewidth]{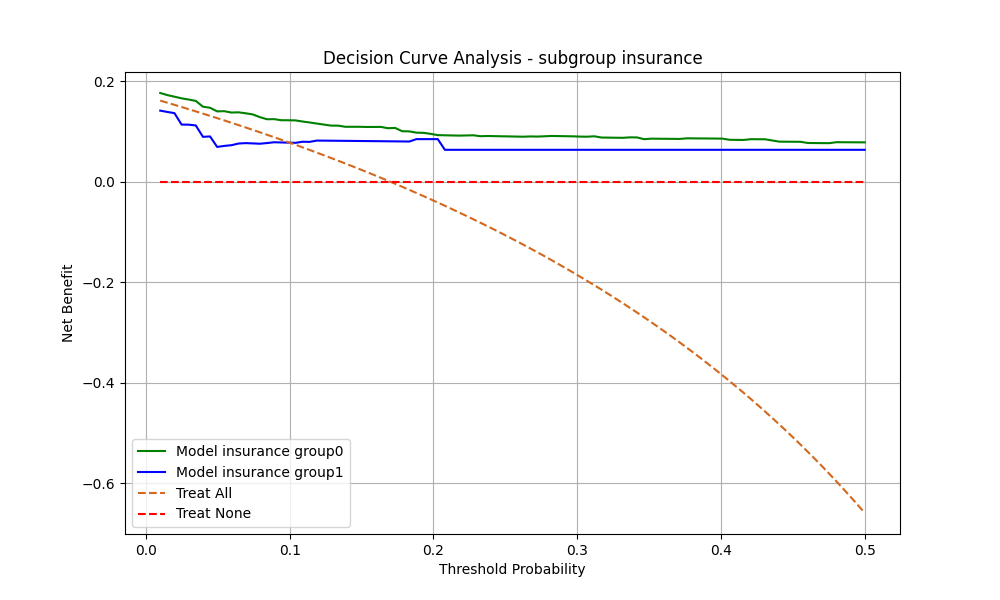}
        \caption{Swin V2 - Insurance (DCA).}
        \label{fig:mBRSET_swin_insu_dca_combined}
    \end{subfigure}

    \vspace{0.5em}

    \begin{subfigure}[t]{0.32\linewidth}
        \centering
        \includegraphics[width=\linewidth]{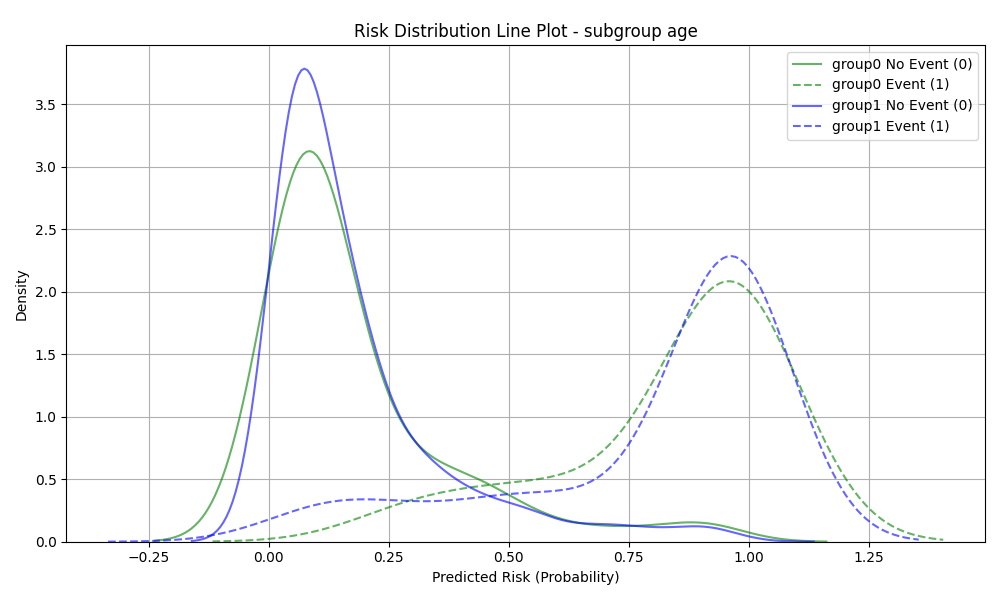}
        \includegraphics[width=\linewidth]{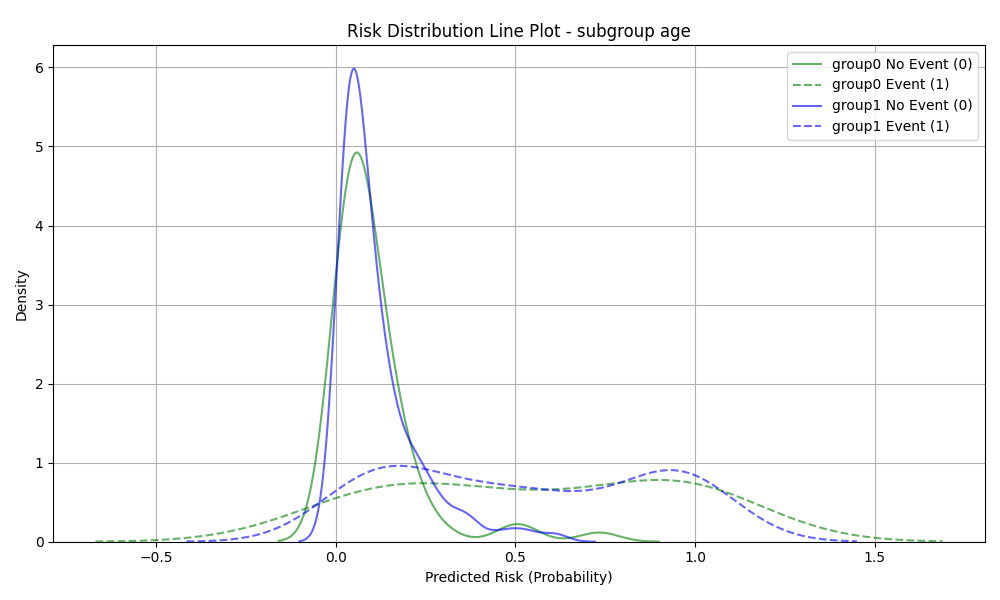}
        \caption{ConvNeXt V2 - Age (Risk Distribution).}
        \label{fig:mBRSET_conv_age_risk_combined}
    \end{subfigure}
    \hfill
    \begin{subfigure}[t]{0.32\linewidth}
        \centering
        \includegraphics[width=\linewidth]{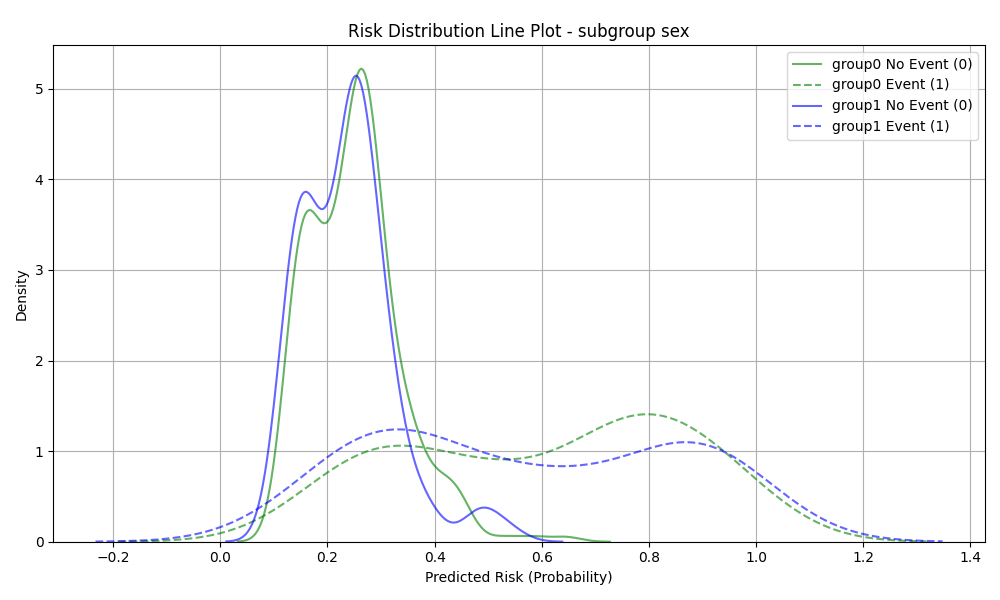}
        \includegraphics[width=\linewidth]{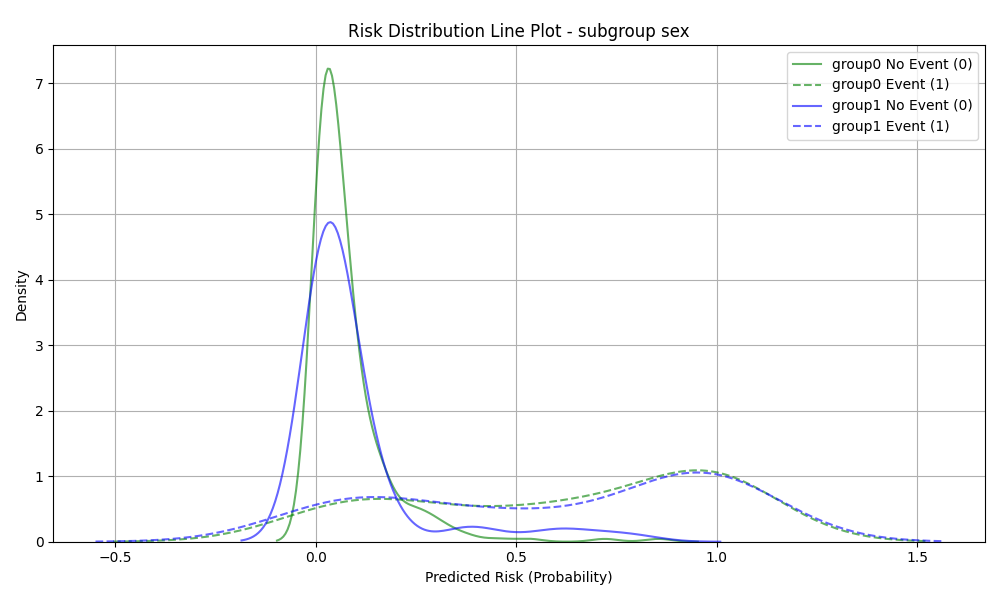}
        \caption{DINOv2 - Sex (Risk Distribution).}
        \label{fig:mBRSET_dino_sex_risk_combined}
    \end{subfigure}
    \hfill
    \begin{subfigure}[t]{0.32\linewidth}
        \centering
        \includegraphics[width=\linewidth]{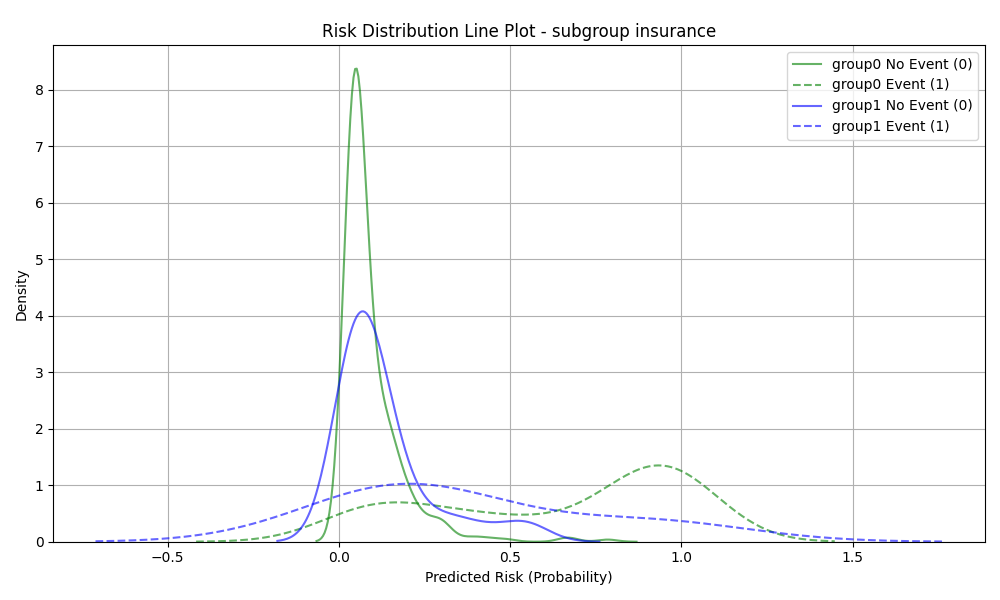}
        \includegraphics[width=\linewidth]{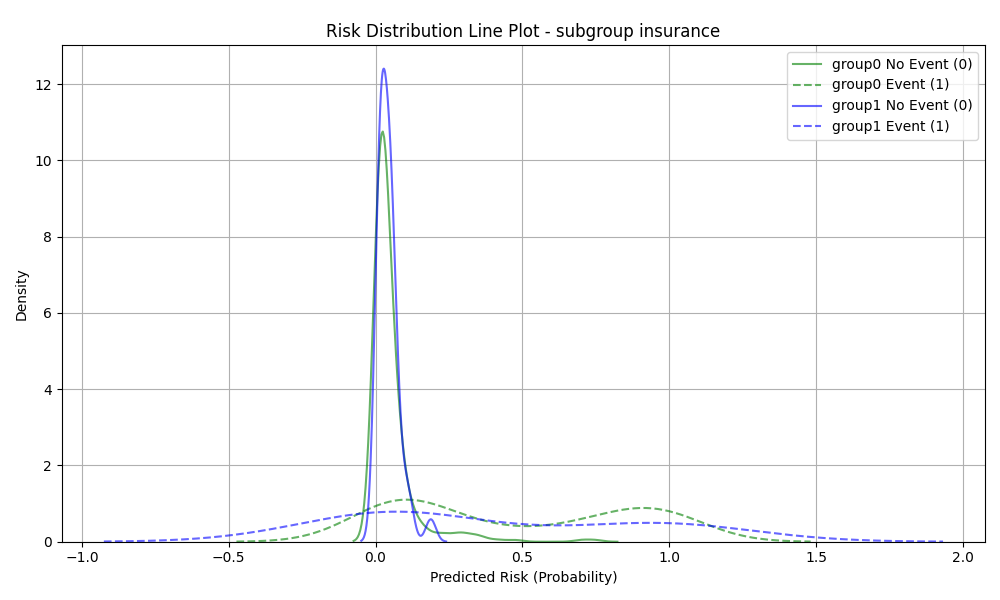}
        \caption{Swin V2 - Insurance (Risk Distribution).}
        \label{fig:mBRSET_swin_insu_risk_combined}
    \end{subfigure}

    \vspace{0.5em}

    \begin{subfigure}[t]{0.32\linewidth}
        \centering
        \includegraphics[width=\linewidth]{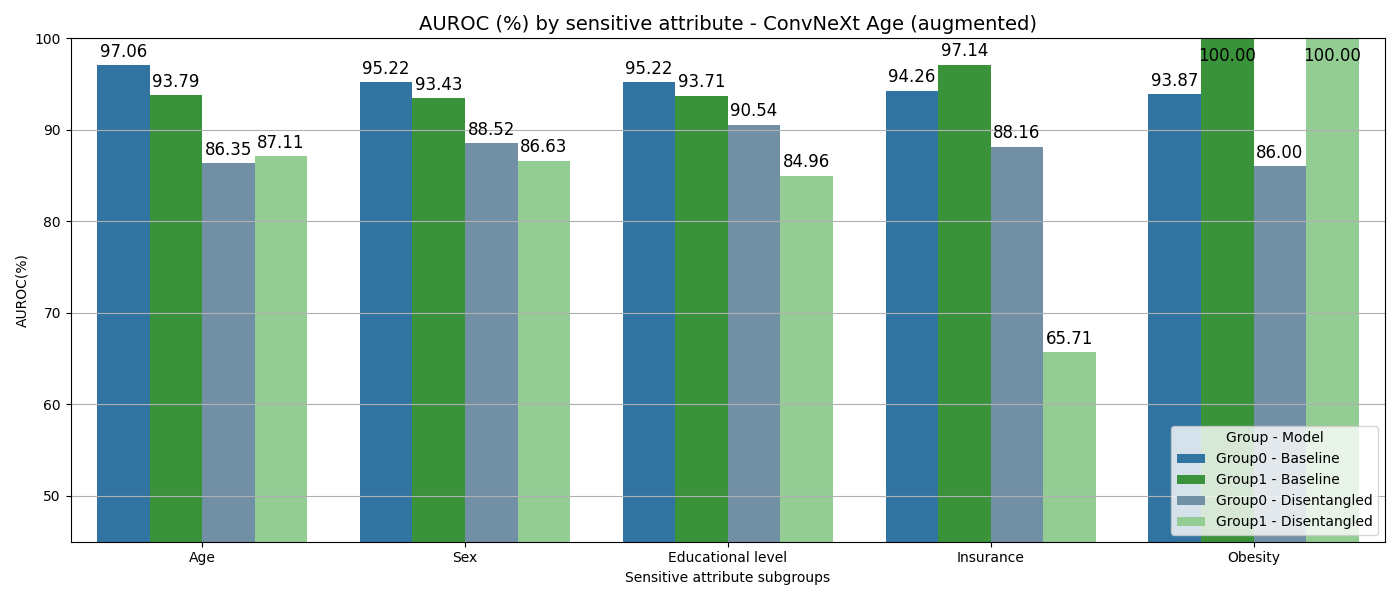}
        \caption{ConvNeXt V2 - Age (AUROC Disparity).}
        \label{fig:mBRSET_conv_age_auroc}
    \end{subfigure}
    \hfill
    \begin{subfigure}[t]{0.32\linewidth}
        \centering
        \includegraphics[width=\linewidth]{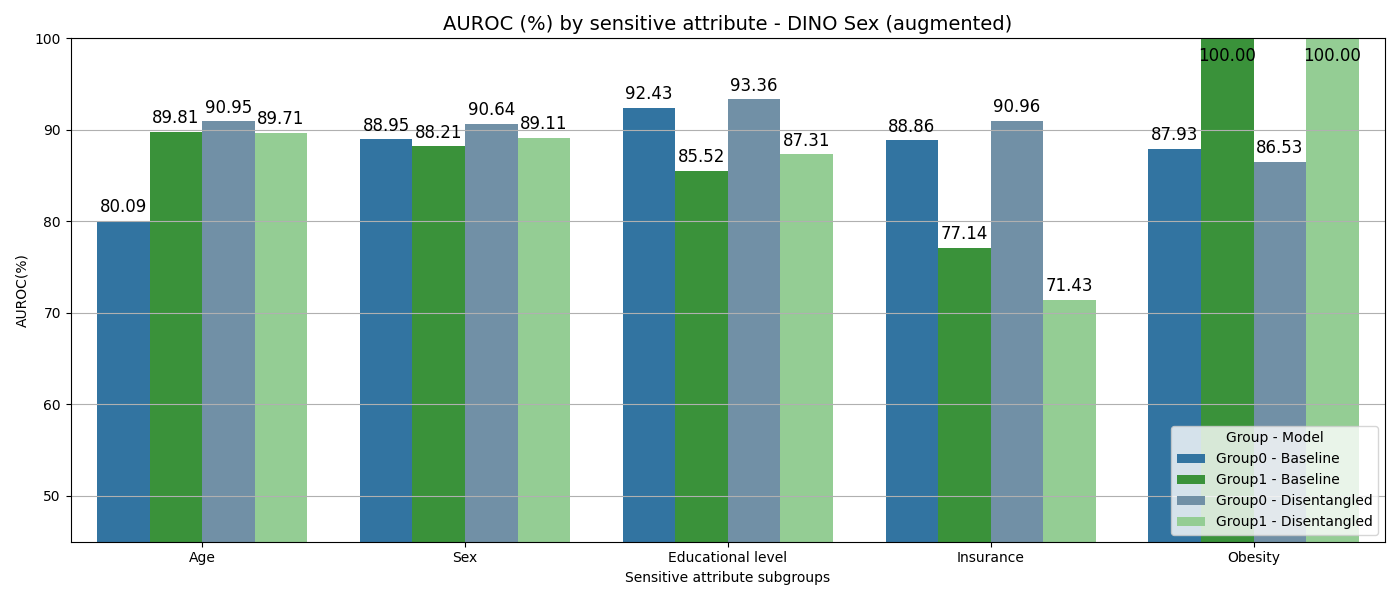}
        \caption{DINOv2 - Sex (AUROC Disparity).}
        \label{fig:mBRSET_dino_sex_auroc}
    \end{subfigure}
    \hfill
    \begin{subfigure}[t]{0.32\linewidth}
        \centering
        \includegraphics[width=\linewidth]{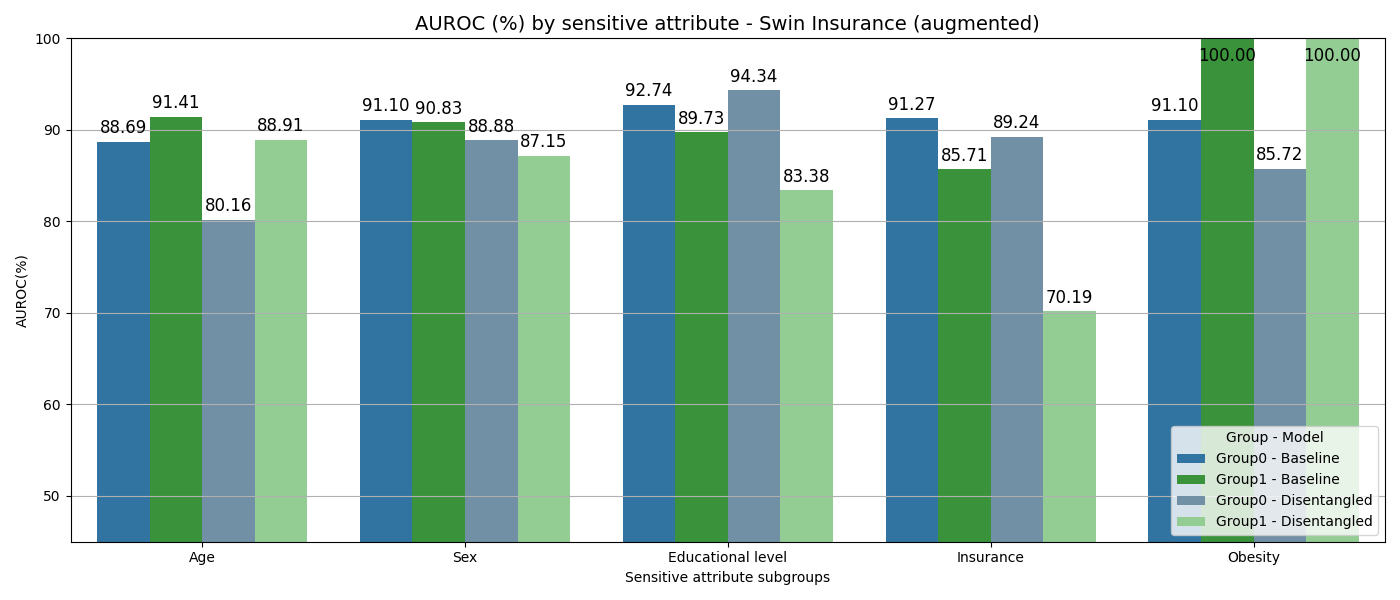}
        \caption{Swin V2 - Insurance (AUROC Disparity).}
        \label{fig:mBRSET_swin_insu_auroc}
    \end{subfigure}

    \caption{Comparison of baseline and disentangled model performance for ConvNeXt V2 (Age), DINOv2 (Sex), and Swin V2 (Insurance) on the mBRSET dataset. Top row: Decision curve analysis comparing baseline (top) and disentangled (bottom) models. Middle row: Risk distribution plots for baseline (top) and disentangled (bottom) models. Bottom row: AUROC disparities across SA groups.}
    \label{fig:mBRSET_combined_all}
\end{figure}

Three representative cases from the mBRSET dataset were selected for analysis based on the highest observed performance disparities and the predictability of SAs. Results are presented in Figure~\ref{fig:mBRSET_combined_all}.


\subsubsection{ConvNeXt V2 - Disentangling medical from age information.}
Disentanglement reduced AUROC from 94\% to 87\%. While age disparity decreased from 4\% to 1\%, nearly all remaining attributes observed increased disparities, most notably insurance (3\% to 21\%). DCA showed no clinical utility gain between age groups. Risk distribution plots revealed lower false positives but higher false negatives, explaining the performance drop. It is possible that the model relied on age-related information for DR prediction, as evidenced by the performance drop when this information was removed and 
its strong age prediction performance. 


\subsubsection{DINOv2 - Disentangling medical from sex information.}
Disentanglement improved AUROC from 88\% to 90\%. Fairness improved for age and educational level (disparities reduced by 8\% and 1\%), while sex and obesity had minimal increases (both by 1\%), and insurance increased by 6\%.
DCA and risk distribution plots revealed improved clinical utility and reduced false positive and false negative rates. The results also suggest that lowering the decision threshold slightly could benefit the model.
Despite weak sex prediction, disentangling sex improved both fairness and generalization.

\subsubsection{Swin V2 - Disentangling medical from insurance information.}
AUROC declined slightly from 91\% to 88\% with disentanglement. Fairness declined across most SAs, including age, insurance, and obesity (5\%, 13\%, and 6\% respectively).
DCA showed a decrease in clinical utility disparity between insurance groups in the disentangled model, while risk distribution plots revealed fewer false positives for insured patients but more false negatives for uninsured patients. 
Disentanglement offered limited fairness improvement and harmed performance. It is unlikely that the model relied on insurance-encoded information for DR prediction, given its weak performance on the insurance prediction task. In these cases, the model appeared to disentangle irrelevant features, focusing on DR-relevant features, even if this did not align perfectly with the targeted SA.

\subsubsection{Summary.}
The fine-grained nature of retinal features makes disentanglement challenging, as it requires removing SA-related information while preserving fine details, which are necessary for accurate clinical interpretation. This trade-off between removing bias and maintaining diagnostic detail complicates the learning process and can hinder model performance.
Furthermore, DR prediction may inadvertently rely on SA-related information, leading to decreased performance when this information is removed.
In cases where confounding factors are present, latent dependencies between SAs and DR-relevant features may persist even after standard disentanglement is applied \cite{ch5:confounding_latents}, limiting effectiveness.


\section{Conclusion and Future Work}
Fairness is crucial in AI healthcare to prevent unequal treatment outcomes and should be assessed before clinical implementation. This work investigated fairness in DR detection and explored disentanglement as a mitigation strategy.
While disentanglement remains a promising direction, its effectiveness in fundus imaging is limited by domain-specific challenges. Bias mitigation in this context is non-trivial, and more robust techniques may be necessary to improve fairness without compromising predictive performance.

\begin{credits}
\subsubsection{\ackname}

\subsubsection{\discintname}
The authors have no competing interests to declare that are relevant to the content of this article.
\end{credits}


%
%
%
\bibliographystyle{splncs04}
\bibliography{mybibliography}

\end{document}